\documentclass{bmvc2k}

\usepackage{booktabs}
\usepackage{amsfonts}
\usepackage{amsmath}
\usepackage{amssymb}
\usepackage{mathtools}
\usepackage{multirow}
\usepackage{algorithm}
\usepackage{algpseudocode}
\usepackage{array}
\usepackage{rotating}
\usepackage{appendix}

\title{Polarimetric Imaging for Perception}

\addauthor{Michael Baltaxe}{michael.baltaxe@gm.com}{1}
\addauthor{Tomer Pe'er}{tomer.1.peer@gm.com}{1}
\addauthor{Dan Levi}{dan.levi@gm.com}{1}

\addinstitution{
 General Motors\\
 Hertzliya, Israel
}

\runninghead{Baltaxe, Pe'er, Levi}{Polarimetric Imaging for Perception}

\graphicspath{ {images/} }

\begin{document}

\maketitle

\begin{abstract}
Autonomous driving and advanced driver-assistance systems rely on a set of sensors and algorithms to perform the appropriate actions and provide alerts as a function of the driving scene. Typically, the sensors include color cameras, radar, lidar and ultrasonic sensors. Strikingly however, although light polarization is a fundamental property of light, it is seldom harnessed for perception tasks. In this work we analyze the potential for improvement in perception tasks when using an RGB-polarimetric camera, as compared to an RGB camera. We examine monocular depth estimation and free space detection during the middle of the day, when polarization is independent of subject heading, and show that a quantifiable improvement can be achieved for both of them using state-of-the-art deep neural networks, with a minimum of architectural changes. We also present a new dataset composed of RGB-polarimetric images, lidar scans, GNSS / IMU readings and free space segmentations that further supports developing perception algorithms that take advantage of light polarization. The dataset can be downloaded \href{https://michaelbaltaxe.github.io/polarimetric_perception}{here}.
\end{abstract}

\section{Introduction}
\label{sec:introduction}
Advanced driver-assistance systems (ADAS) and autonomous vehicles need to interpret the surrounding environment to plan and act appropriately on the road. To do so, modern vehicles are equipped with a set of sensors and algorithms that carry out a variety of perception tasks such as free space detection \cite{sne_roadseg, plard, stixelnet}, lane detection \cite{3d_lanenet, towards_end_to_end_lane_detection}, object detection \cite{yolo, ssd, faster_rcnn}, 3D pose estimation \cite{single_stage_3d_object_detection, graph_rnn_3d_object_detection}, depth estimation \cite{monodepth_v2, dorn}, etc.

The quality of the perception output is a function of the quality of the sensor suite installed in the vehicle. Currently, RGB cameras, ultrasonic sensors and radars are standard equipment in production vehicles, and in the near future lidars will also be readily available thanks to the steady decrease in their size and price.

RGB-polarimetric cameras are sensors that measure light polarization, in addition to light intensity and color. These cameras are already in use in several industrial applications; for example, to detect defects in surfaces and to improve image quality by removing reflections. However, to the best of our knowledge, they have not been used in the automotive domain. In this work we explore the potential of using RGB-polarimetric cameras to improve the performance of algorithms employed in common autonomous driving perception tasks.

Since light polarization is a complex function of light properties, scene properties and viewing direction, we limited this study to the case when the sun was high in the sky (around midday). This way the polarization state of collected light was independent of vehicle heading, and consistent readings were made in all driving directions.

We focus on free space detection and depth estimation. Both are of paramount importance for automated driving and ADAS systems, but are also useful for viewing systems when creating surround and bowl-views of the environment.

The purpose of free space detection is to segment the input image such that all pixels where the vehicle can drive are labeled  `1' and all the other pixels in the image are labeled `0', thus creating a mask indicating the ``drivable'' space which enables planing and navigation.

By contrast, depth estimation calculates the depth from the camera sensor to the imaged object in each pixel, thus yielding a dense distance map of the scene. There are currently several methods for depth estimation, many of which use either multiple view geometry or active imaging. Here, we examined monocular depth estimation (monodepth), where the depth is estimated from a single image taken with a passive camera (i.e., without active illumination). In our case, we used polarimetric information in addition to the RGB image to estimate depth.

Note that specialized hardware is used in many applications to achieve depth estimation. Lidar, for example, has been the prime choice to enrich the sensing suite and achieve high quality reliable sensing. Compared to this solution, the RGB-polarimetric camera has the desirable properties of being cheaper and providing a dense map that is readily aligned with the RGB image, without the need for complex alignment procedures.

The development and evaluation of perception algorithms calls for a dataset that supports the target tasks. Modern perception methods typically use convolutional neural networks (CNN) \cite{lecun_cnn} and other deep learning methods \cite{lecun_bengio_hinton_deep_learning}, which require a large amount of data for training. Although there are several available public-domain automotive datasets \cite{kitti, kitti360, waymo, nuscenes, bdd100k, cityscapes, blin_dataset}, few have polarimetric data. To the best of our knowledge, the only automotive dataset that includes polarimetric information is \cite{blin_dataset}. Its major drawback is that the RGB and polarimetric images were obtained from different devices, so that the synchronization between modalities is only partial. To overcome this hurdle, we built a dataset of RGB-polarimetric data composed of RGB images, polarimetric images (angle of linear projection and degree of linear polarization), lidar point clouds, GPS and inertial measurements.

This dataset is composed of 12,627 images from 6 different locations. The data were collected around noon in fair weather. The intrinsic parameters of the camera (focal distance and principal point) and extrinsic parameters (translation and rotation) between all the hardware components were calibrated, thus providing full alignment between the camera and all other sensing elements.
Note that alignment between the RGB and polarimetric images was easy to achieve since a single sensor was used to create both images (see section \ref{sec:method dataset}).

The contribution of this work is twofold. First, it presents a dataset of RGB-polarimetric data with naturalistic driving scenarios useful for several perception tasks. Then, we present a detailed analysis of the performance of key perception algorithms using RGB-polarimetric data which are compared to the performance obtained when only RGB data are available.

The remainder of this paper is organized as follows.
Section \ref{sec:related work} presents related work concerning perception in the automotive domain and the use of light polarization for perception. Section \ref{sec:method} presents the specifics of the dataset and the approach used to include polarimetric data in two perception tasks. Section \ref{sec:experiments} reports the experiments and section \ref{sec:conclusions} concludes.

\section{Related Work}
\label{sec:related work}

\subsection{Shape from Polarization}
\label{sec:related work shape from polarization}
Shape from polarization aims to recreate surface shapes from light polarization measurements. The method presented in \cite{recovery_from_diffuse_polar} was one of the first to recover surface normals from polarimetric images of objects with diffusive-reflecting materials using analytical methods. Newer methods such as \cite{deep_sfp, sfp_in_the_wild} use deep neural networks to cope with the fact that real world objects exhibit specular and diffusive behavior. The method in \cite{p2d} used a CNN to estimate depth from polarimetric and grayscale images rather than estimating surface normals.

\subsection{Monocular Depth Estimation}
\label{sec:related work monocular depth estimation}
Monocular depth estimation is crucial to computer vision. Supervised methods such as in \cite{dorn, semisupervised_monodepth} were trained to learn a direct distance metric for each pixel from ground truth data collected with specialized depth sensors. By contrast, self-supervised methods such as \cite{monodepth, monodepth_v2, sfm_learner, packnet} are much more data efficient because they take advantage of the geometric relationships within a scene when the camera moves in space. For this purpose, pairs of consecutive frames are used to learn pose and depth estimation networks, while minimizing the reprojection error. 

\subsection{Free Space Detection}
\label{sec:related work free space detection}
Free space detection has been widely studied by the autonomous driving community. In this task, the system outputs a segmentation of the environment in which the vehicle can drive, usually corresponding to a road. Several approaches exclusively use camera information, while others also use lidar point clouds. Early work introduced in \cite{stixelnet} developed a method that takes an RGB image as input and uses a CNN to extract \textit{stixels}, a compact representation of free space. More recent works such as \cite{sne_roadseg, sne_roadseg_plus, usnet} use lidar point clouds to extract surface normals which are fed along with the RGB image into a fully convolutional network. Similarly, \cite{plard} used a two-stream neural network to process an RGB image together with an \textit{altitude difference} image extracted from the lidar point cloud. Yet another approach was taken by \cite{bevformer}, which used multiple cameras and a vision transformer to yield a precise segmentation. 

\subsection{Datasets for Driving Perception Tasks}
\label{sec:related work datasets}
There are several specialized open datasets for driving perception tasks. Probably the best known are described in \cite{kitti, kitti360, nuscenes, waymo, bdd100k, cityscapes}. All these datasets include RGB images, and several also provide lidar point clouds. The annotation level covers 2D object bounding boxes, 3D object bounding boxes, drivable area delineation, object tracking, instance segmentation and segmantic segmentation for the image or point cloud modalities.
In \cite{blin_dataset}, a dataset of RGB and polarimetric images was used for object detection. In this dataset, two different cameras were used to capture the two modalities but no extrinsic calibration was calculated, so that the RGB and polarization image pairs were not perfectly aligned or synchronized.


\section{Method}
\label{sec:method}

\subsection{Dataset}
\label{sec:method dataset}
Since our perception methods rely on deep learning techniques, we needed a large dataset with polarimetric information.
We built a custom setup to gather data for our experiments. The setup
included the following hardware:
\begin{enumerate}
	\item RGB-polarimetric camera (Lucid Vision TRI050S-QC with Sony IMX250MYR CMOS color sensor).
	\item Lidar (Velodyne Alpha Prime).
	\item GNSS / INS (OxTS RT3000).
\end{enumerate}

The RGB-polarimetric camera outputs RGB values with polarization filters at four different angles, from which intensity (I), angle of linear polarization (AoLP) and degree of linear polarization (DoLP) images can be calculated, as explained below. The resolution of the restored images was 1.25 megapixels, with a field of view of $60^\circ$.

Let $P_{0}, P_{45}, P_{90}, P_{135}$ be the intensity of the polarization images obtained by the camera, where the subscripts indicate the orientation angle of the polarization filter. Then, following standard practice \cite{sfp_in_the_wild, p2d}, we calculated the intensity, AoLP and DoLP as follows:
\begin{align}
	I &= \frac{\left( P_0 + P_{45} + P_{90} +  E_{135} \right)}{2}  \\
	DoLP &= \frac{\sqrt{\left( P_0 - P_{90} \right)^2 + \left(P_{45} - P{135}\right)^2}}{I} \\
	AoLP &= \frac{1}{2} \arctan\left({\frac{P_{45} - P_{135}}{P_0 - P_{90}}}\right).
\end{align}


The lidar camera system was synchronized temporally with the lidar used as the trigger for the camera, providing a trigger signal each time the revolving head reaches the $0^\circ$ mark. The camera and lidar data were collected at a frame rate of 10 Hz, and the GNSS / INS was sampled at 100 Hz.

The camera's intrinsic parameters were calibrated using the standard chessboard method implemented in OpenCV \cite{opencv_library}. The extrinsic parameters between the camera and the lidar (translation and rotation) were calibrated by calculating the rigid transformation that achieved the smallest distance between several planes in the three main directions (in terms of least squares), extracted independently with the lidar and the camera. In this case, the camera planes were extracted by capturing images of a chessboard.

Figure \ref{fig:data samples} presents a few examples of the collected data. The cyclic color coding in the AoLP image shifts from red for $0^\circ$ to magenta for $179^\circ$. In the DoLP image, 0 corresponds to black and 1 corresponds to yellow.

\begingroup
\setlength{\tabcolsep}{3pt} 
\begin{figure*}
	\centering
	\begin{tabular}{cccc}
		\includegraphics[width=0.23\textwidth]{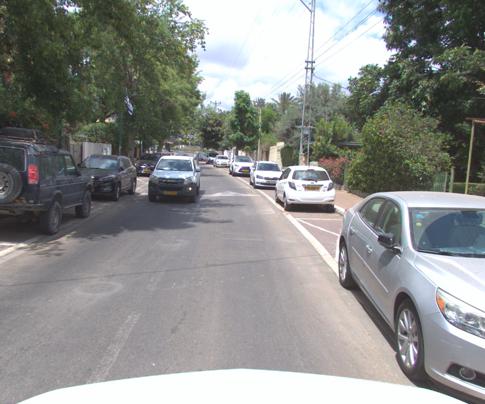} & \includegraphics[width=0.23\textwidth]{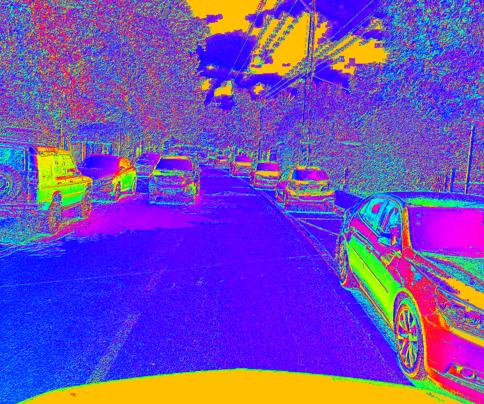} & \includegraphics[width=0.23\textwidth]{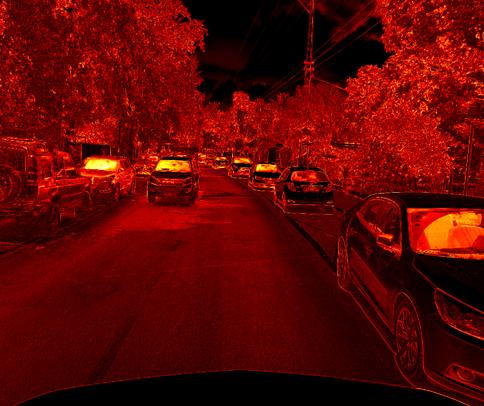} & \includegraphics[width=0.23\textwidth]{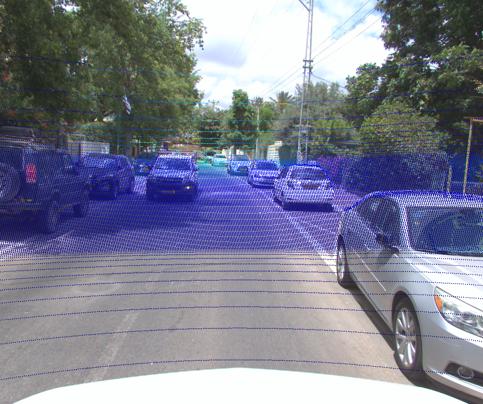} \\
		\includegraphics[width=0.23\textwidth]{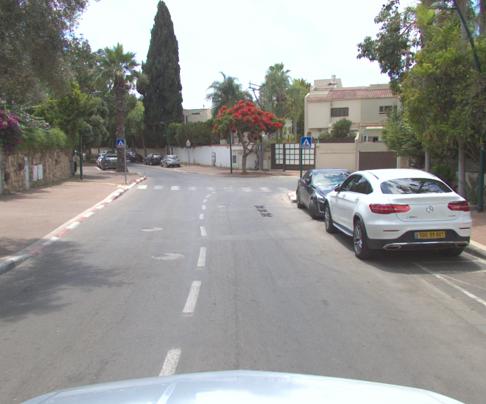} & \includegraphics[width=0.23\textwidth]{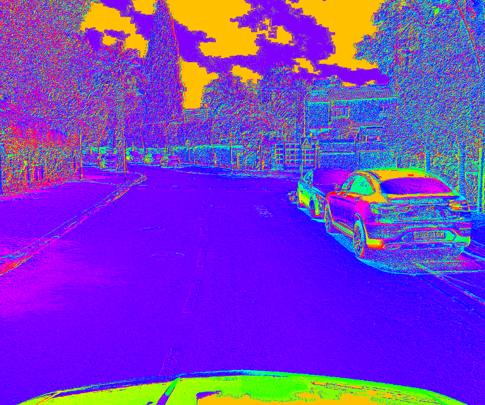} & \includegraphics[width=0.23\textwidth]{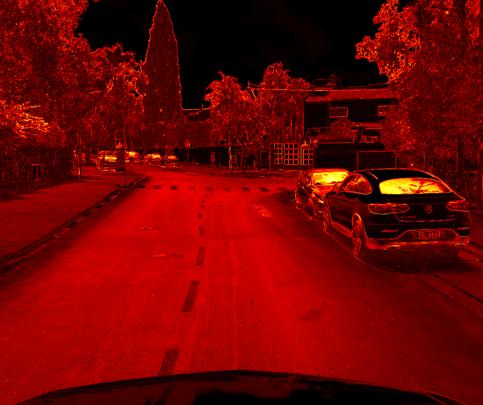} & \includegraphics[width=0.23\textwidth]{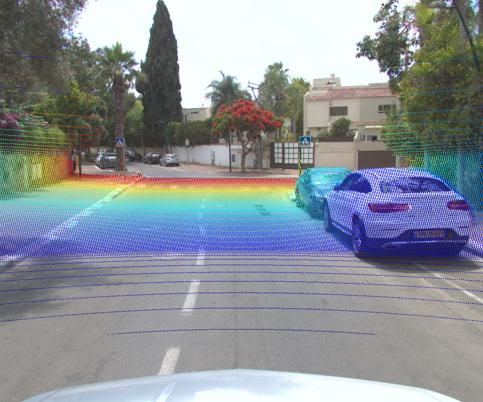} \\
		\includegraphics[width=0.23\textwidth]{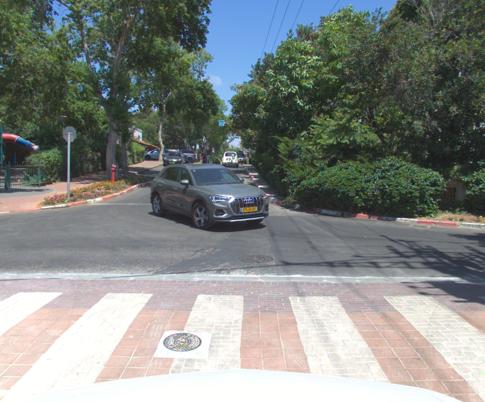} & \includegraphics[width=0.23\textwidth]{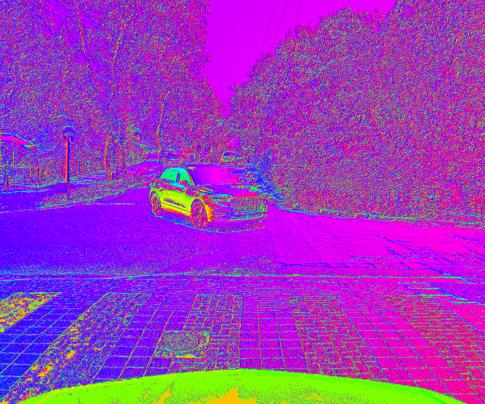} & \includegraphics[width=0.23\textwidth]{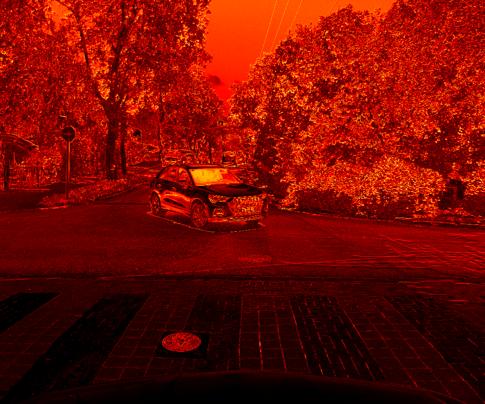} & \includegraphics[width=0.23\textwidth]{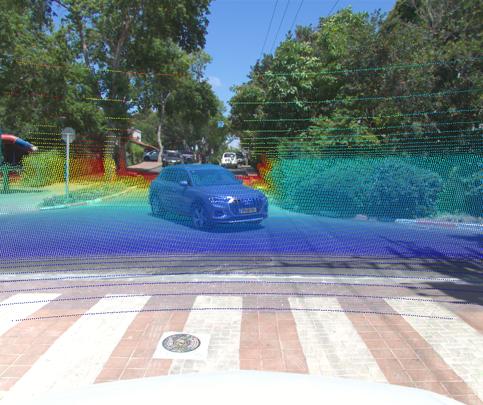}
	\end{tabular}
	\caption[Examples of collected data]{Examples of collected data. Each row shows a different sample with RGB (left), AoLP (middle left), DoLP (middle right) and lidar projected on RGB (right). The cyclic color map in the AoLP images goes from red for $0^\circ$ to magenta for $179^\circ$. In the DoLP images black corresponds to $0$ and yellow to $1$.}
	\label{fig:data samples}
\end{figure*}
\endgroup

Windshields tend to light up in the DoLP image because they have a high degree of linear polarization due to the smoothness of the glass. In addition, the road and other horizontal elements tend to have an AoLP close to $0^\circ$ (purple) because their normal is aligned upwards and the electric field of light oscillates perpendicularly. Note as well that the AoLP depends not only on geometry, but also on the material, as shown for example on the side of the two vehicles in the second row. While both vehicles have the same orientation with their sides located vertically, for the black car the AoLP is close to $90^\circ$ (green), but for the white car the AoLP is close to $0^\circ$ (purple). This effect can also be seen on the side window of the white car in the second row, which has the same geometry as the car side, but is made of different material.

The dataset was composed of 12,627 images from 6 different locations. These locations represent typical suburbs where the scenes are not remarkably cluttered, but provide a good distribution of vehicles, pedestrians and buildings. No highways were included in the dataset since we used self-supervised methods for monodepth estimation (section \ref{sec:method monocualar depth estimation}), which are known to degrade strongly when the dataset contains vehicles that move at speeds similar to the ego-vehicle's speed. Additionally, the dataset includes free space segmentation of 8,141 images. The segmentations were created in a semi-automatic way using the SAM segmentation method \cite{sam} followed by manual refinement.

\subsection{Free Space Detection}
\label{sec:method free space}

Our objective in this study was to quantify the potential benefits of using an RGB-polarimetric camera for perception tasks. For the free space scenario, we based our method on the SNE-RoadSeg architecture \cite{sne_roadseg}, one of the top-scoring methods in the KITTI road benchmark with open-source code. The original network takes an RGB image and a depth image (usually acquired with a lidar) as inputs, and outputs a free space segmentation. This method initially estimates the surface normal from the depth image using the SNE module, and then uses a deep neural network to perform the final segmentation. The SNE-RoadSeg network has two streams: one extracts features from the surface normal channels and the other extracts features from a concatenation of the RGB and the surface normal images.

Surface normals are tightly correlated to polarimetric measurements. Specifically, following \cite{recovery_from_diffuse_polar} (under the assumption that the materials are not ferromagnetic), the specular AoLP ($\phi_{s}$) and DoLP ($\rho_{s}$) and the diffusive AoLP ($\phi_{d}$) and DoLP ($\rho_{d}$) are related to the surface normal's azimuth ($\alpha$) and zenith ($\theta$) angles as follows:
\begin{align}
	\phi_s &= \alpha - \frac{\pi}{2}
	\label{eq:specular AoLP form azimuth} \\
	\rho_s &= \frac{2 \sin^2(\theta)\cos(\theta)\sqrt{n^2 - \sin^2(\theta)}}{n^2 - \sin^2(\theta) - n^2\sin^2(\theta) + 2\sin^4(\theta)}
	\label{eq:specular DoLP form zenith} \\
	\phi_d &= \alpha \\
	\rho_d &= \frac{\left( n - \frac{1}{n} \right)^2 \sin^2(\theta)}{2 + 2n^2 - \left( n + \frac{1}{n} \right)^2 \sin^2(\theta) + 4 \cos(\theta) \sqrt{n^2 - \sin^2(\theta)}}
\end{align}
where $n$ is the refractive index of the material of the object being imaged.

We hypothesized that a network that is able to extract information from the surface normal should also infer successfully from polarimetric data, when trained properly. For this reason, our architecture was exactly the same as the SNE-RoadSeg, except that the SNE module was dropped and the input surface normal channels were replaced by a concatenation of polarimetric channels as follows:
\begin{equation}
	P = \left[ \sin(2 \cdot AoLP), \cos(2 \cdot AoLP), 2 \cdot DoLP - 1 \right].
	\label{eq:polarimetric features}
\end{equation}

We used the sine and cosine functions on the AoLP to cope with the fact that the AoLP is a cyclic function, where a measurement of $0^\circ$ is equivalent to $180^\circ$. We scaled the DoLP to be in the range [-1, 1], as for the other two features.

\begin{figure*}
	\centering
	\includegraphics[width=0.75\textwidth]{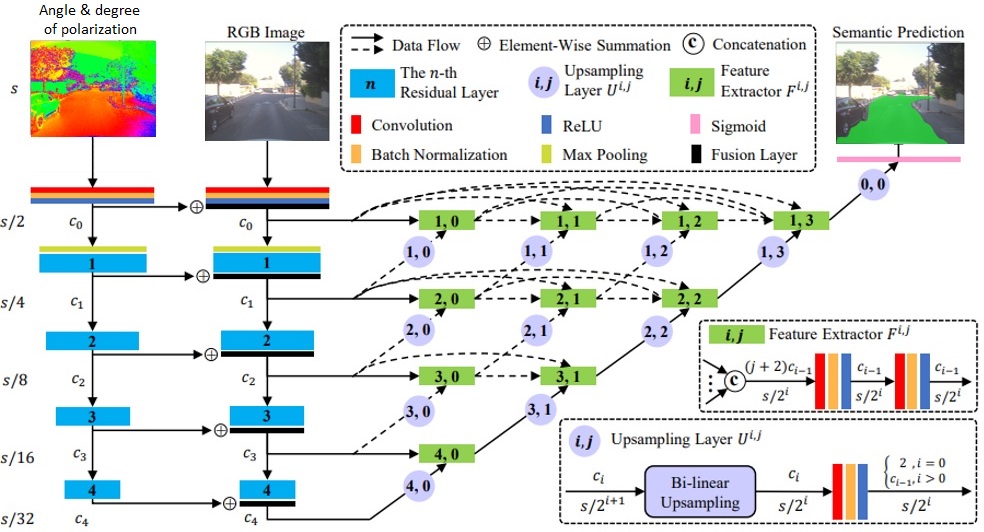}
	\caption[RGB-Polar-RoadSeg architecture]{RGBP-RoadSeg architecture.\footnotemark}	
	\label{fig:polar sne roadseg}
\end{figure*}

The architecture used for the free space detection network, dubbed \textit{RGBP-RoadSeg}, is depicted in Figure \ref{fig:polar sne roadseg}.

\footnotetext{Image adapted from \cite{sne_roadseg} with permission from the authors.}

\subsection{Monocular Depth Estimation}
\label{sec:method monocualar depth estimation}

For the monocular depth estimation we drew on the well-known monodepth v2 framework \cite{monodepth_v2}. The main idea is to take pairs of consecutive frames over time and learn two networks: one for depth estimation and one for estimating the relative camera pose between the two frames. A loss is calculated by warping the estimated depth from one frame to the next by applying the estimated camera pose and projecting back to the image. This is a clever self-supervised method to learn depth, its main strength is that no manual labeling is needed.

The original work used RGB images as input to the networks. In our system, we used the RGB image concatenated with the three polarimetric features described in equation \ref{eq:polarimetric features} as input. The architecture is the same as the one used in monodepth v2.

\section{Experiments}
\label{sec:experiments}

\subsection{Free Space Detection}
\label{sec:experiments free space detection}

\noindent
\textbf{Data:}\quad
We used the SAM system \cite{sam} to create automatic segmentations of the scenes by providing as input prompt a point right in front of the vehicle's hood, which can be expected to be part of the road with high probability. Then, the segmentations were inspected and manually refined. Overall we extracted 8,141 segmentations which were divided into train (6,206 images), validation (856 images) and test (969 images) splits. The train, test and validation data were mutually exclusive geographically. The number of images used in this task was smaller than the full dataset as extracting segmentations is quite expensive.

\medskip
\noindent
\textbf{Evaluated Methods:}\quad
Our \textit{RGBP-RoadSeg} as described in section \ref{sec:method free space} is compared to other RoadNet incarnations. The closest RGB-only implementation, named \textit{RGB-RoadSeg}, was exactly like the RGBP-RoadSeg, but the left stream in Figure \ref{fig:polar sne roadseg} was completely dropped, leaving only the RGB input and the skip connections of the original network. The \textit{P-RoadSeg} droped the left stream of the network, and used the polarimetric features of equation \ref{eq:polarimetric features} as input to the right stream. Finally, we evaluated the standard \textit{SNE-RoadSeg} nework \cite{sne_roadseg} which used as input RGB images and depth images processed by the SNE module, allowing to compare lidar-based and polarization-based methods.

\medskip
\noindent
\textbf{Metrics:}\quad
We used the standard metrics of \cite{kitti_road}: accuracy, precision, recall, maximum F-score ($\mathrm{F_{max}}$) and average precision (AP). Intersection over union (IoU) was also evaluated. As in \cite{kitti_road}, all metrics were calculated on bird's-eye view projections of the scenes.

\medskip
\noindent
\textbf{Results:}\quad
The results of the free space detection are presented in Table \ref{tab:free space detection}.
First, note that P-RoadSeg yielded mediocre results, implying that polarization alone does not carry enough information for this task. RGB-RoadSeg provided much better results, which tells us that the RGB modality is more suited for free space estimation. The best results, however, were obtained by RGBP-RoadSeg which uses both RGB and polarimetic information suggesting that both modalities are complementary and carry independent information.
RGBP-RoadSeg (polarimetric camera) is on a par with SNE-RoadSeg (lidar), although it is important to recall the noon-time constraint on the polarimetric camera.

\begin{table}
	\centering
	\begin{tabular}{ccccccc}
		\toprule
		Method & Accuracy & Precision & Recall & $\mathrm{F_{max}}$ & IoU & AP \\
		\midrule
		RGB-RoadSeg & 0.979 & 0.949 & 0.968 & 0.953 & 0.902 & 0.974 \\
		P-RoadSeg & 0.865 & 0.845 & 0.534 & 0.641 & 0.467 & 0.634 \\
		RGBP-RoadSeg & \textbf{0.986} & \textbf{0.966} & \textbf{0.972} & \textbf{0.968} & \textbf{0.939} & \textbf{0.994} \\
		\midrule
		SNE-RoadSeg & 0.985 & 0.967 & 0.967 & 0.965 & 0.934 & 0.993 \\
		\bottomrule	
	\end{tabular}
	\caption{Results for the free space detection task.}
	\label{tab:free space detection}
\end{table}

\begin{figure}
	\centering
	\begingroup
	\setlength{\tabcolsep}{0pt} 
	\renewcommand{\arraystretch}{0} 
	\begin{tabular}{ccccc}
		\rotatebox{90}{\footnotesize \textcolor{white}{tt} RGB-RSeg} &
		\includegraphics[width=0.24\linewidth]{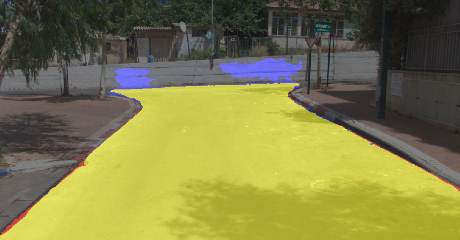} &
		\includegraphics[width=0.24\textwidth]{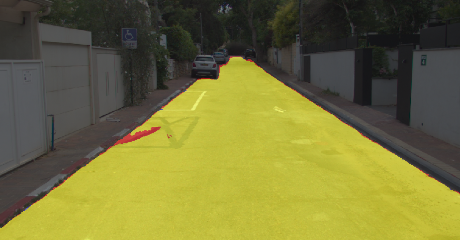} &
		\includegraphics[width=0.24\linewidth]{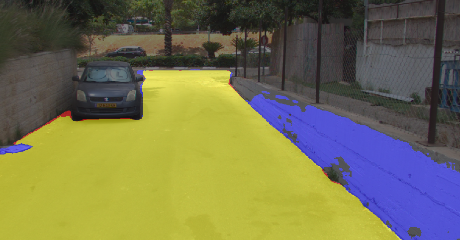} &
		\includegraphics[width=0.24\linewidth]{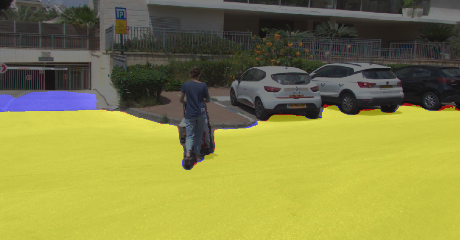} 
		\\
		\rotatebox{90}{\footnotesize \textcolor{white}{t} RGBP-RSeg} &
		\includegraphics[width=0.24\linewidth]{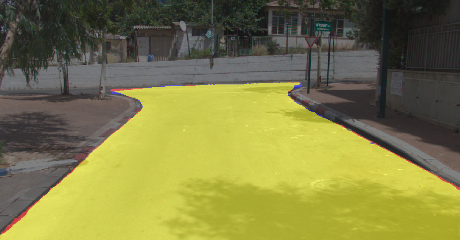} &
		\includegraphics[width=0.24\textwidth]{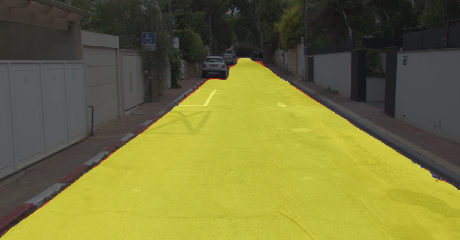} &
		\includegraphics[width=0.24\linewidth]{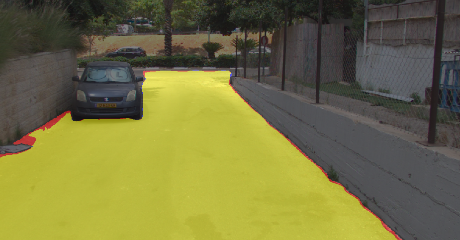} &
		\includegraphics[width=0.24\linewidth]{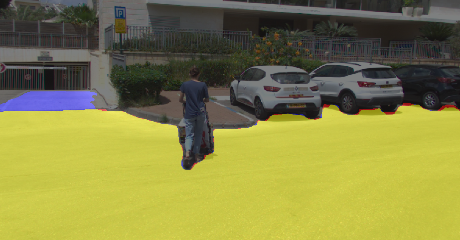} 
		\\
		\rotatebox{90}{\footnotesize \textcolor{white}{tt} SNE-RSeg} &
		\includegraphics[width=0.24\linewidth]{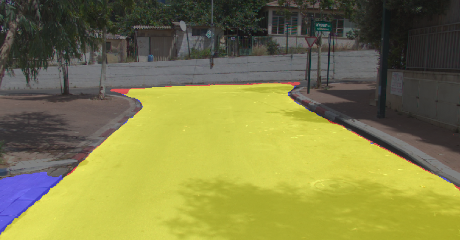} &
		\includegraphics[width=0.24\textwidth]{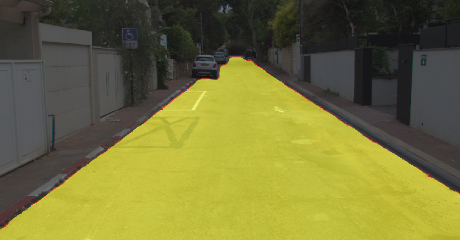} &
		\includegraphics[width=0.24\linewidth]{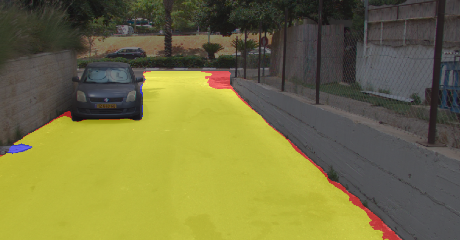} &
		\includegraphics[width=0.24\linewidth]{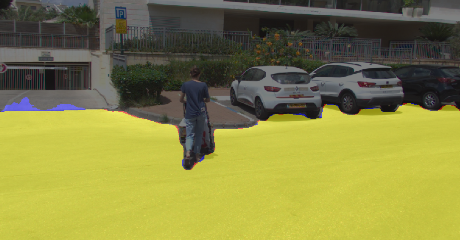}
	\end{tabular}
	\endgroup
	\caption{Qualitative results for  free space detection. Yellow, blue and red correspond to true positive, false positive and false negative respectively. RGB-RoadSeg bled in low contrast regions and missed painted road areas. SNE-RoadSeg relies on depth and bled to ground outside of the road. The right column is a failure case: semantics are needed to find the edge.}
	\label{fig:free space detection qualitative results}
\end{figure}

Figure \ref{fig:free space detection qualitative results} shows some qualitative results. Note the extent to which low contrast areas were improved by the use of polarization data. This makes sense since color contrast is not always correlated with polarization contrast. For example, the wall and road in the third column have similar colors (yielding poor color contrast), but the wall orientation has a $90^\circ$ shift with respect to the road, which yields a high contrast in the AoLP image.

\subsection{Monocular Depth Estimation}
\label{sec:experiments monocular depth estimation}

\noindent
\textbf{Data:}\quad
We divided the dataset into train (6,116 images), validation (778 images) and test (778 images) splits. The number of images used does not include the full dataset since the self-supervised monocular depth paradigm cannot use frames where the vehicle is static. We set a minimum speed of 15 Km/h and use all frames where the vehicle moves at higher speed. The train, validation and test data are divided so that there is no geographical overlapping.

\medskip
\noindent
\textbf{Evaluated Methods:}\quad
Our baseline method, denoted \textit{RGB-Depth}, simply performs monocular depth estimation using RGB images, as is commonly done. Then, we drop the RGB images and instead use the polarimetric features in equation \ref{eq:polarimetric features} as input to the system, we call this \textit{P-Depth}. Next, we analyze the posibility to use RGB and polarimetric data in a synergistic manner. In this case, we stack the RGB images with the polarimetric features in equation \ref{eq:polarimetric features} and use this as input to the monodepth method, we refer to this setup as \textit{RGBP-Depth}. Finally, we pre-train the model on RGB images and fine tune it using the stacking of RGB and polarimetric features (the polarimetric features are initialized randomly). This last method is regarded as \textit{pt-RGBP-Depth}.

\medskip
\noindent
\textbf{Metrics:}\quad
We used the standard metrics introduced in \cite{eigen_depth} to quantify both the error and the accuracy of the methods. For details, see \cite{eigen_depth}.

\medskip
\noindent
\textbf{Results:}\quad
The results of these experiments are presented in Table \ref{tab:monodepth estimation}. First of all, note that the RGB results are consistent with the results of the original paper, showing that our dataset is relevant for the task. Using polairmetric data instead of RGB we see an improvement, this is probably since the polarimetric modality includes a lot of information that is relevant to the task of depth estimation. Stacking together the RGB and polarimetric data in the RGBP-Depth method yields further improvement, showing that both modalities are complementary and do not carry the same information. Finally, the pt-RGBP-Depth which pre-trains on RGB and uses both modalities for fine tuning reaches the best results. Qualitative results are presented in Figure \ref{fig:depth estimation qualitative results}. RGBP-Depth shows sharper edges and fuller structures.

\begingroup
\setlength{\tabcolsep}{3pt} 
\begin{table*}
	\centering
	\begin{tabular}{ccccccccc}
		\toprule
		\multirow{3}{*}{Method} & \multicolumn{4}{c}{Error metric $\downarrow$} && \multicolumn{3}{c}{Accuracy metric $\uparrow$} \\
		\cmidrule{2-5} \cmidrule{7-9} 
		& Abs Rel & Sq Rel & RMSE & RMSE Log && $\delta < 1.25$ & $\delta < 1.25^2$ & $\delta < 1.25^3$ \\
		\midrule
		RGB-Depth & 0.094  &   0.838  &   6.389  &   0.166  &&   0.904  &   0.964  &   0.984 \\
		P-Depth &   0.091  &   0.811  &   6.325  &   0.164  &&   0.907  &   0.966  &   0.985  \\
		RGBP-Depth & 0.089  &   0.770  &   6.172  &   0.161  &&   0.911  &   \textbf{0.968}  &   \textbf{0.986} \\
		pt-RGBP-Depth &   \textbf{0.086}  &   \textbf{0.767}  &   \textbf{6.109}  &   \textbf{0.158}  &&   \textbf{0.915}  &   \textbf{0.968}  &   0.985  \\
		\bottomrule
	\end{tabular}
	\caption{Results for the monodepth estimation task.}
	\label{tab:monodepth estimation}
\end{table*}
\endgroup

\begin{figure}
	\centering
	\begingroup
	\setlength{\tabcolsep}{0pt} 
	\renewcommand{\arraystretch}{0} 
	\begin{tabular}{cccc}
		& RGB-Depth & P-Depth & pt-RGBP-Depth \\
		\includegraphics[width=0.24\linewidth]{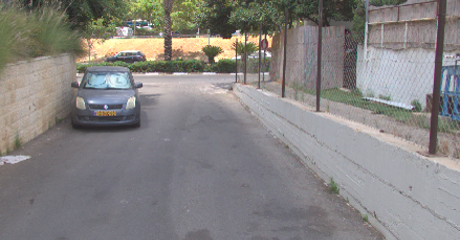} &
		\includegraphics[width=0.24\linewidth]{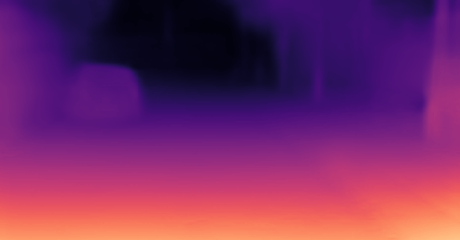} &
		\includegraphics[width=0.24\textwidth]{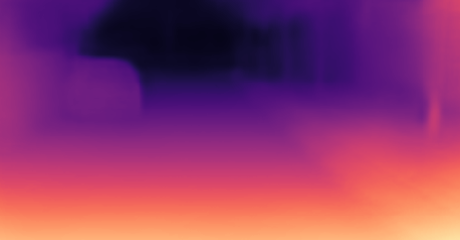} &
		\includegraphics[width=0.24\linewidth]{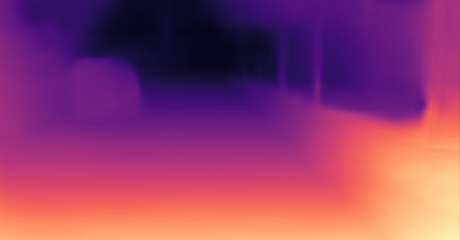} 
		\\
		\includegraphics[width=0.24\linewidth]{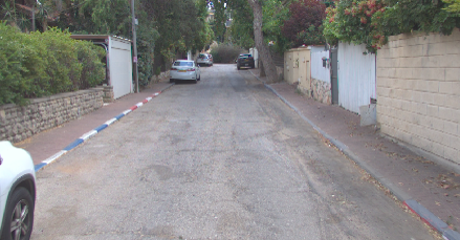} &
		\includegraphics[width=0.24\linewidth]{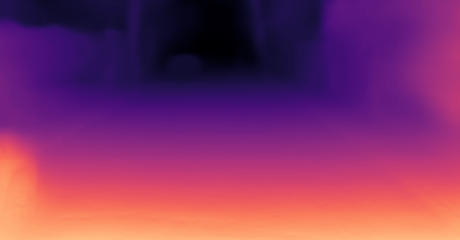} &
		\includegraphics[width=0.24\textwidth]{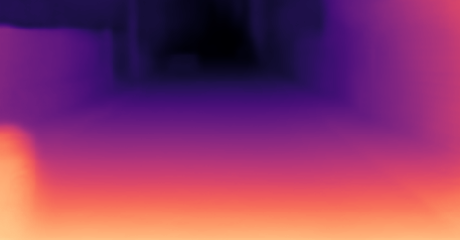} &
		\includegraphics[width=0.24\linewidth]{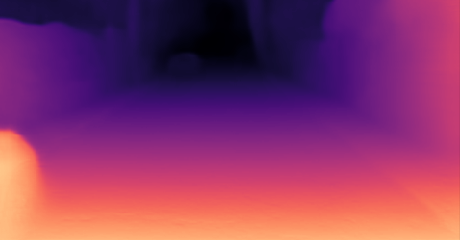}
		\\
		\includegraphics[width=0.24\linewidth]{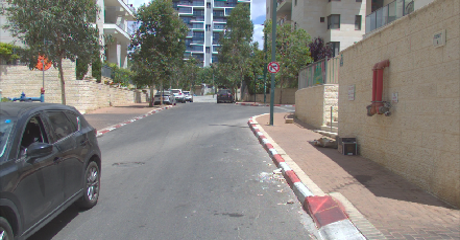} &
		\includegraphics[width=0.24\linewidth]{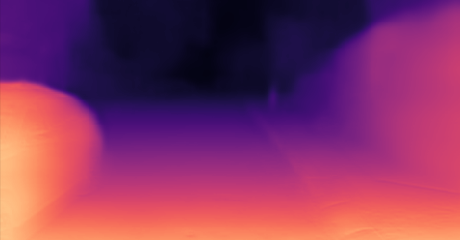} &
		\includegraphics[width=0.24\textwidth]{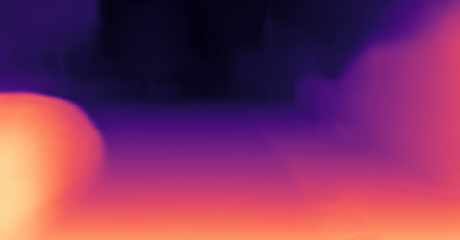} &
		\includegraphics[width=0.24\linewidth]{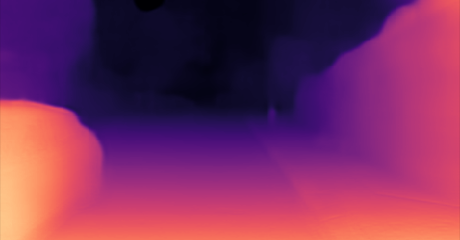}
	\end{tabular}
	\endgroup
	\caption{Qualitative results for the depth estimation task. pt-RGBP-Depth yields sharper edges and better recovers all structures.}
	\label{fig:depth estimation qualitative results}
\end{figure}

\section{Conclusion and Future Work}
\label{sec:conclusions}

In this work we examined the advantages of polarimetric imaging. We analyzed the extent to which two perception tasks can be improved when polarization information is used along with standard RGB images.

Our data collection methodology consisted of an RGB-polarimetric camera, a lidar and a GNSS / IMU system. We showed that this setup makes it possible to gather a large database that can serve many perception tasks since all the modalities are aligned and synchronized.

Our evaluation of the free space and monocular depth estimation showed that by using RGB and polarization information we could improve the results as compared to using RGB information alone. Interestingly, this improvement was achieved with only minor architectural changes. 


Future work will focus on extending the models to cope with situations where the noon-time constraint does not hold.

\section*{Acknowledgments}
We thank Shaked Magali, Tal Piterman, Tony Eyal Naim, Eldar Riklis and Gilad Oshri for their help building the data acquisition setup. We also thank Tzvi Philipp, Noa Garnett and Emanuel Mordechai for their useful ideas.

\bibliography{refs}

\newpage
\appendix
\flushleft {\huge \textbf{\textcolor{bmv@sectioncolor}{Appendices}}}

\section{Implementation Details}
Here we present implementation details of the models used.
\label{sec:implementation details}

\subsection{Free Space Detection}
\label{sec:implementation details free space detection}
All the Road-Seg networks \cite{sne_roadseg} were trained with an initial learning rate of 0.05. The learning rate was reduced by a ratio of 0.9 (as in the the original SNE-RoadSeg code) every 10 epochs. The training batch size was 8 and the loss used was binary cross entropy. The networks were trained until the evaluation metrics stopped improving on the validation set, which usually occurred after about 100 epochs. The architecture of the two streams of the Road-Seg network was a ResNet-152 with all weights initialized randomly to avoid biases coming from pre-training with large RGB datasets (like ImageNet \cite{imagenet}), compared to the smaller size of the polarimetric data.

\subsection{Monocular Depth Estimation}
\label{sec:implementation details monocular depth estimation}
All the monodepth v2 models \cite{monodepth_v2} were trained with the standard parameters provided by the official code. That is, the learning rate was 1e-4 and the batch size was 12. The depth and pose encoders used were ResNet-18 networks. The networks were trained until the evaluation metrics stopped improving on the validation set, usually after about 60 epochs.

\section{Additional Qualitative Results}
\label{sec:additional qualitative results}

\begin{figure}
	\centering
	\begingroup
	\setlength{\tabcolsep}{0pt} 
	\renewcommand{\arraystretch}{0} 
	\begin{tabular}{ccc}
		RGB-RoadSeg & RGBP-RoadSeg & SNE-RoadSeg \\
		\includegraphics[width=0.32\textwidth]{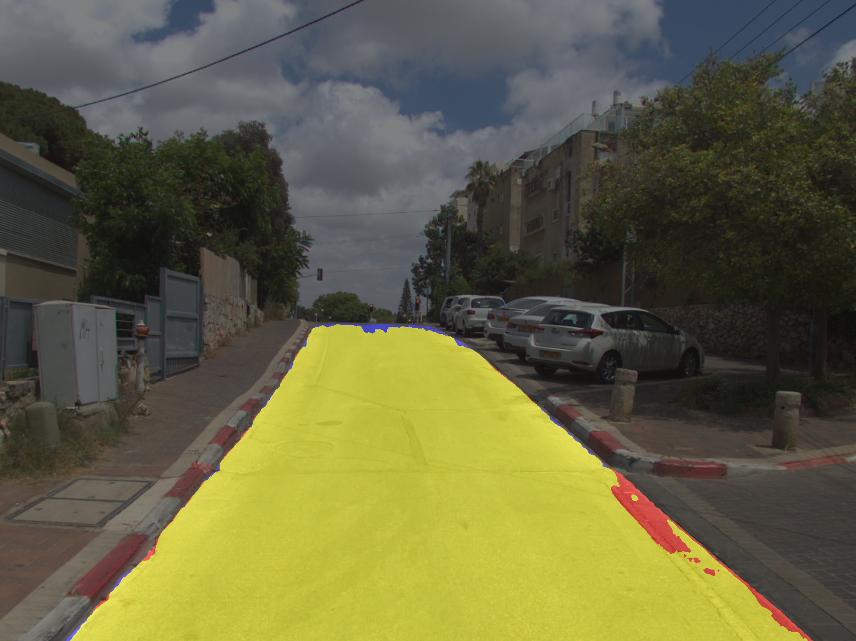} &
		\includegraphics[width=0.32\textwidth]{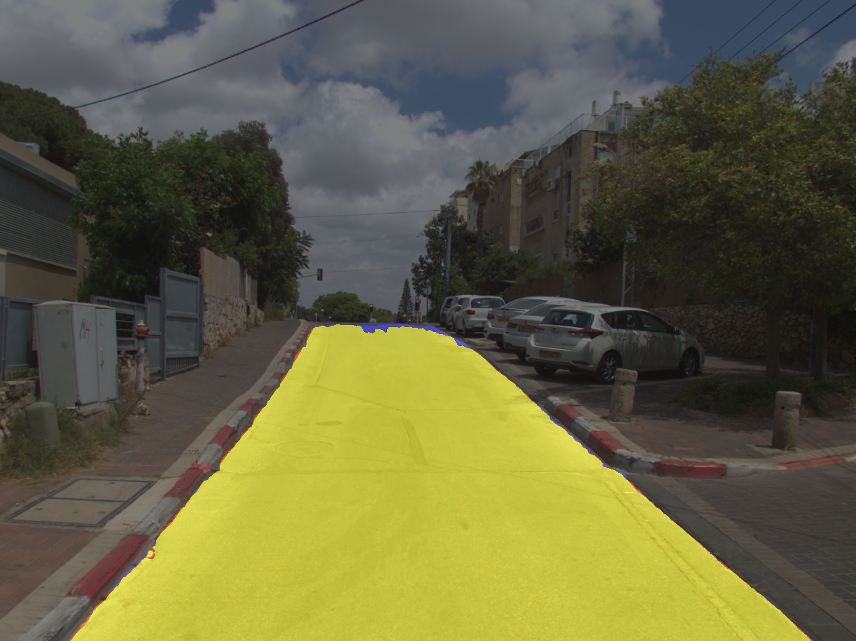} &
		\includegraphics[width=0.32\textwidth]{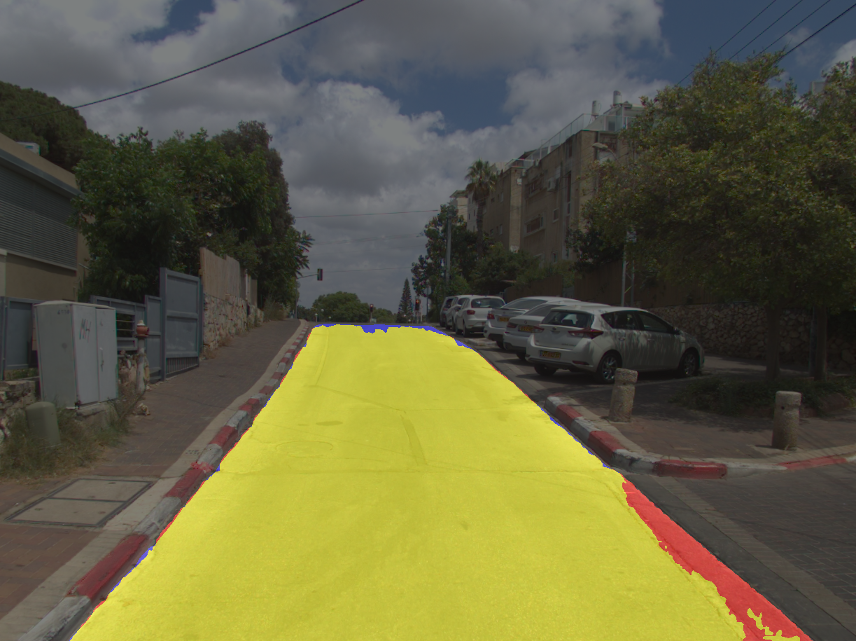}
		\\
		\includegraphics[width=0.32\textwidth]{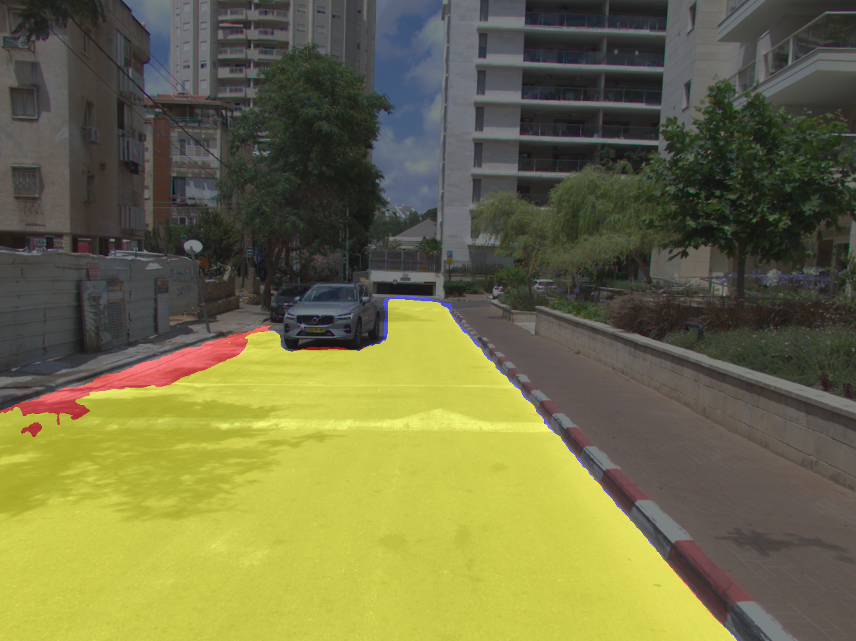} &
		\includegraphics[width=0.32\textwidth]{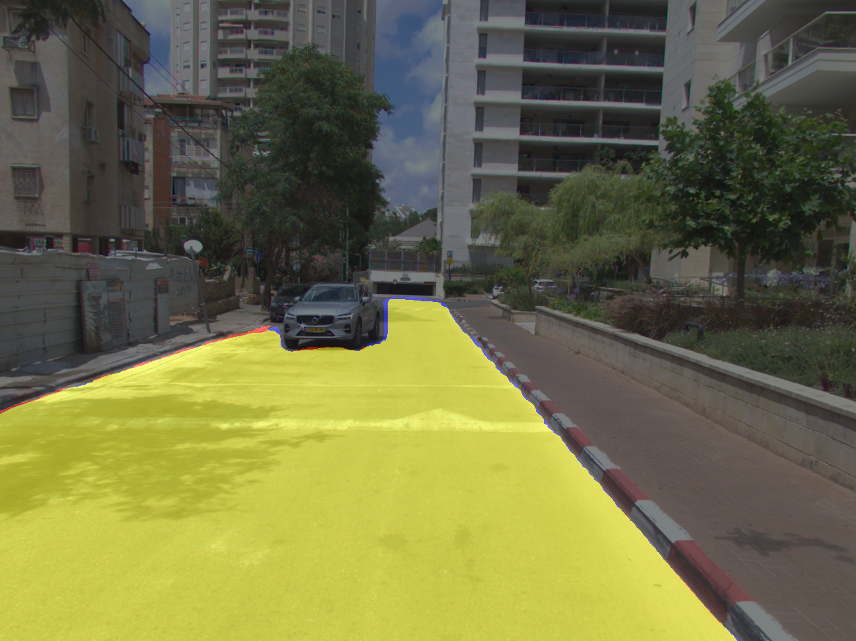} &
		\includegraphics[width=0.32\textwidth]{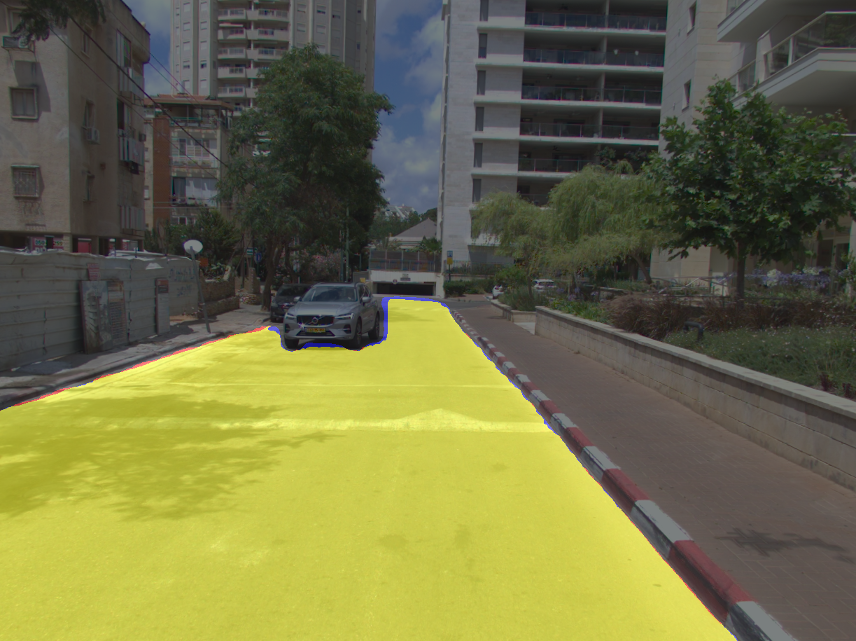}
		\\
		\includegraphics[width=0.32\textwidth]{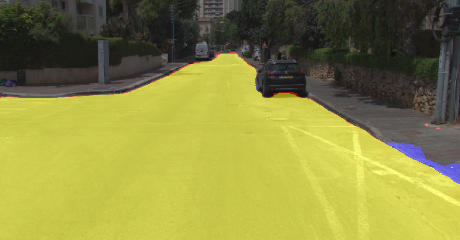} &
		\includegraphics[width=0.32\textwidth]{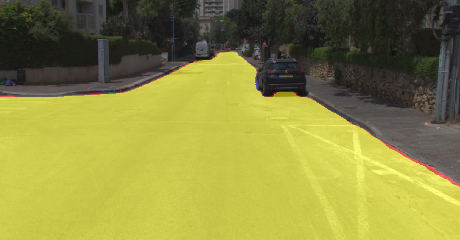} &
		\includegraphics[width=0.32\textwidth]{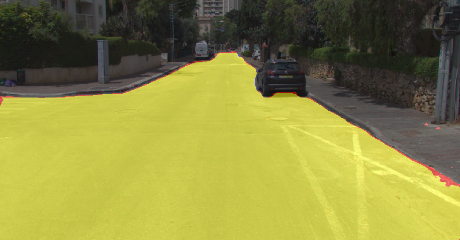}
		\\
		\includegraphics[width=0.32\textwidth]{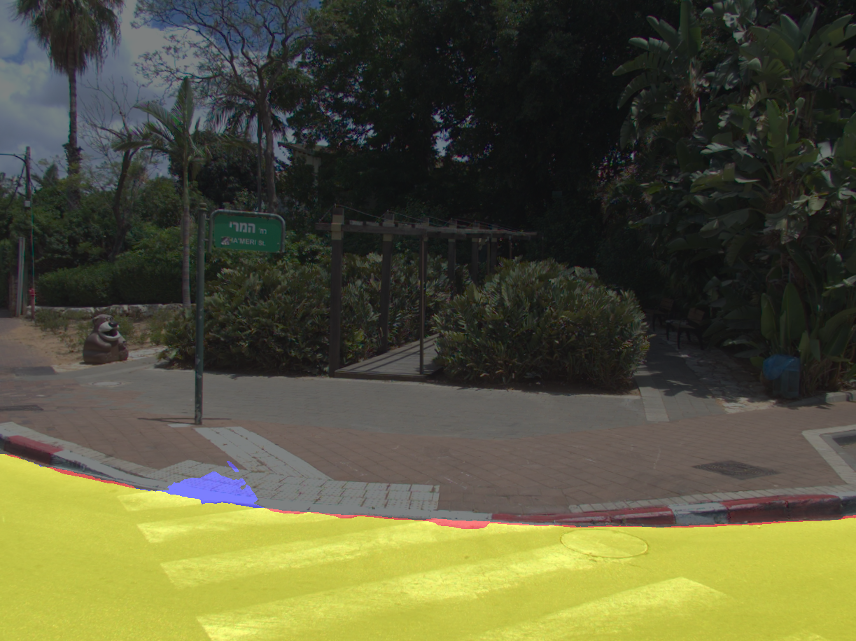} &
		\includegraphics[width=0.32\textwidth]{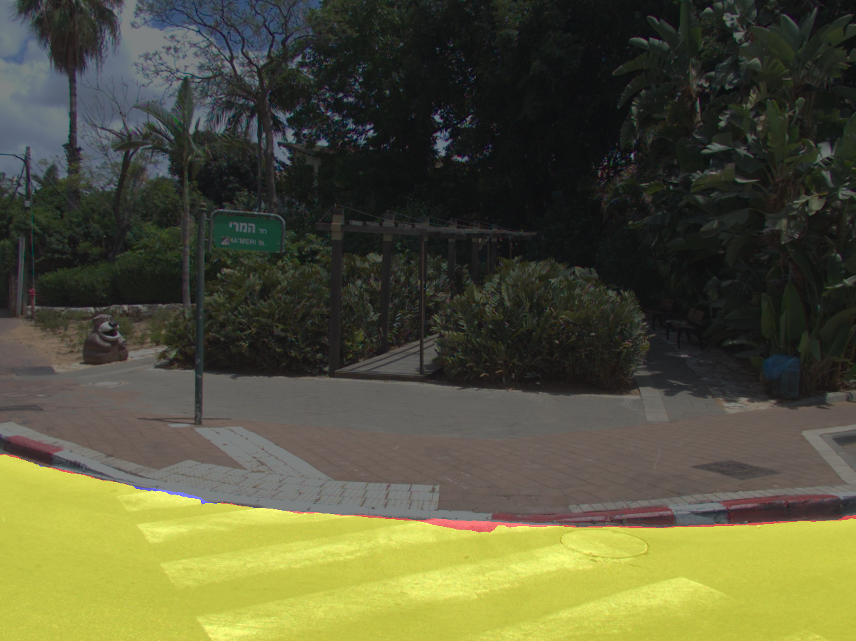} &
		\includegraphics[width=0.32\textwidth]{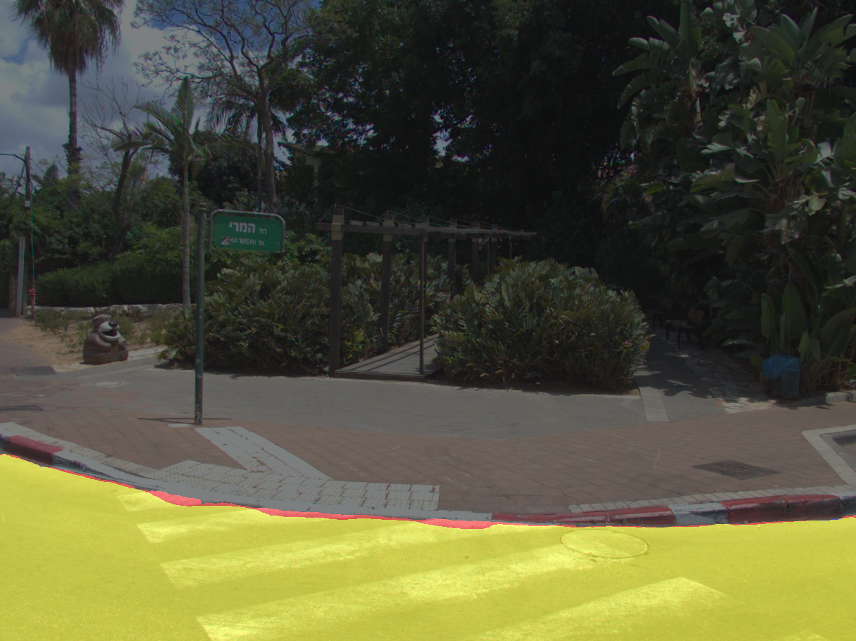}
		\\
		\includegraphics[width=0.32\textwidth]{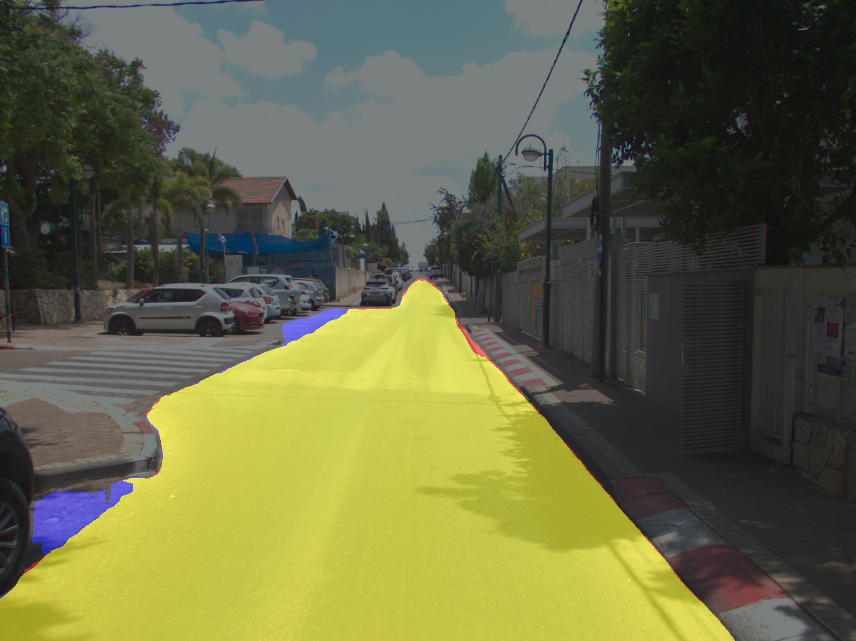} &
		\includegraphics[width=0.32\textwidth]{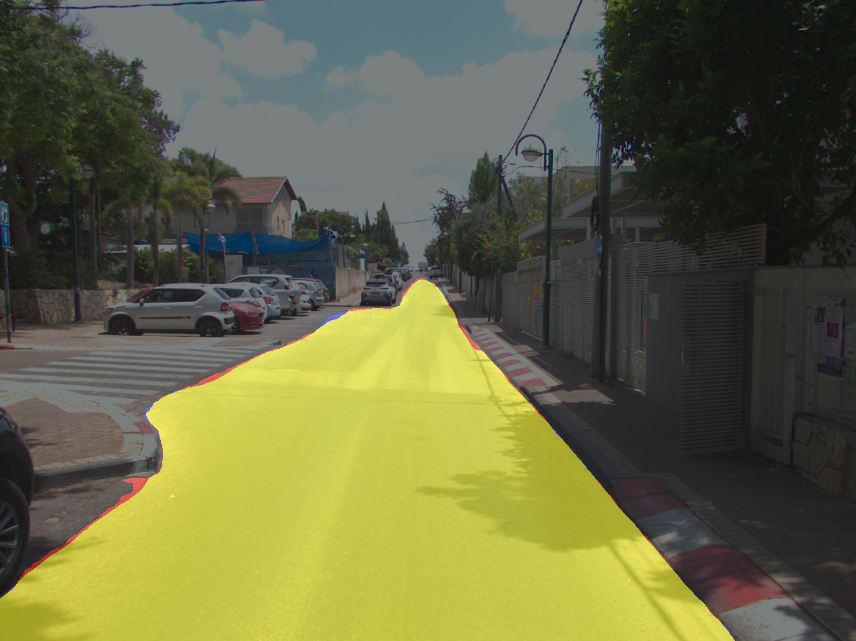} &
		\includegraphics[width=0.32\textwidth]{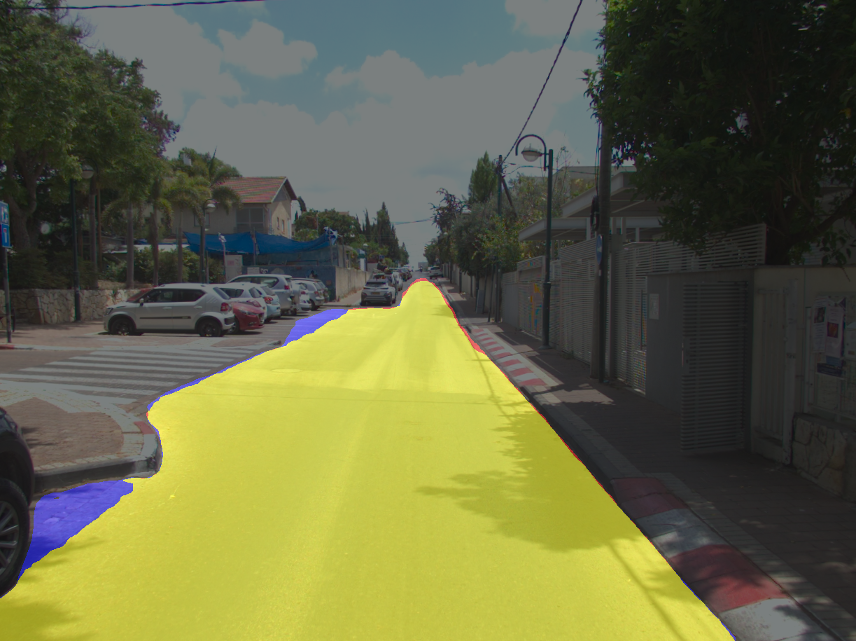}
		\\
		\includegraphics[width=0.32\linewidth]{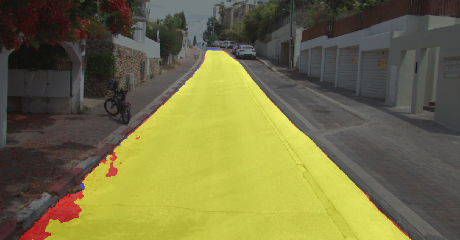} &
		\includegraphics[width=0.32\linewidth]{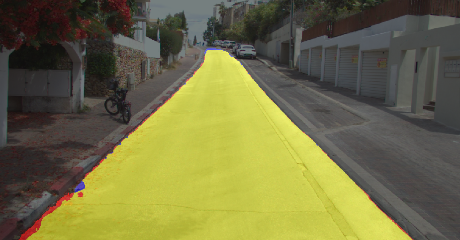} &
		\includegraphics[width=0.32\linewidth]{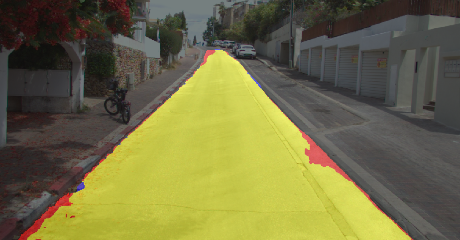}
		\\
	\end{tabular}
	\endgroup
	\caption{Qualitative results for  free space detection. Yellow, blue and red correspond to true positive, false positive and false negative respectively. Best viewed zoomed in.}
	\label{fig:free space detection qualitative results appendix}
\end{figure}

\begin{figure}
	\centering
	\begingroup
	\setlength{\tabcolsep}{0pt} 
	\renewcommand{\arraystretch}{0} 
	\begin{tabular}{cccc}
		& RGB-Depth & P-Depth & pt-RGBP-Depth \\
		\includegraphics[width=0.24\linewidth]{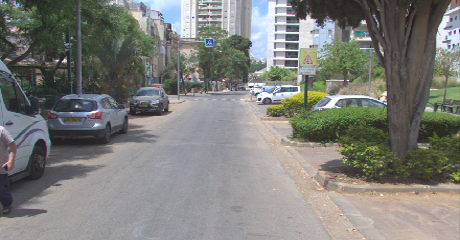} &
		\includegraphics[width=0.24\linewidth]{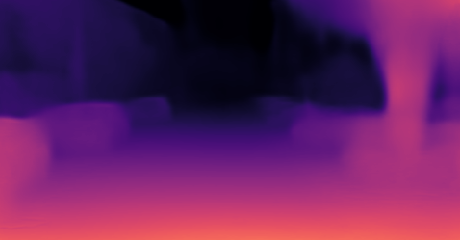} &
		\includegraphics[width=0.24\linewidth]{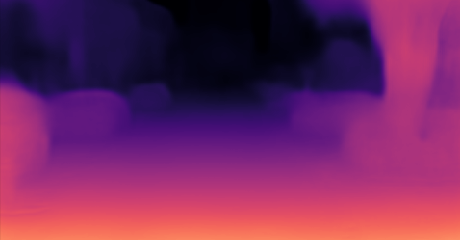} &
		\includegraphics[width=0.24\linewidth]{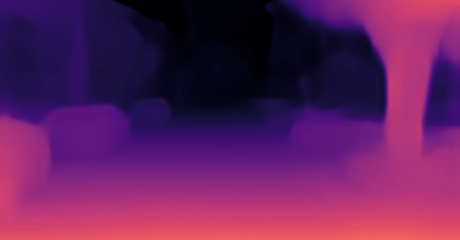} 
		\\
		\includegraphics[width=0.24\linewidth]{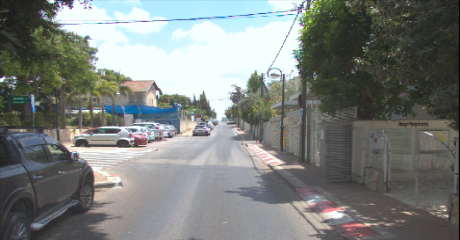} &
		\includegraphics[width=0.24\linewidth]{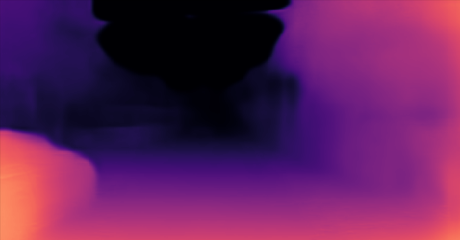} &
		\includegraphics[width=0.24\textwidth]{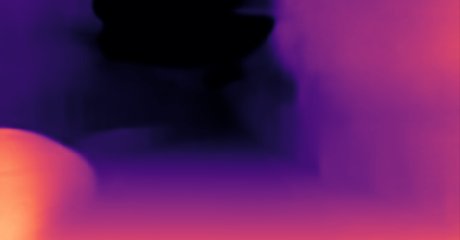} &
		\includegraphics[width=0.24\linewidth]{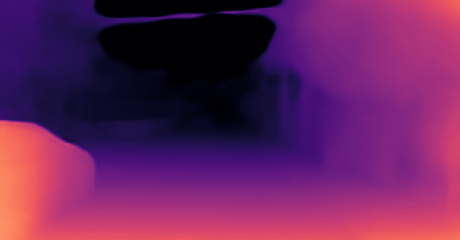}
		\\
		\includegraphics[width=0.24\linewidth]{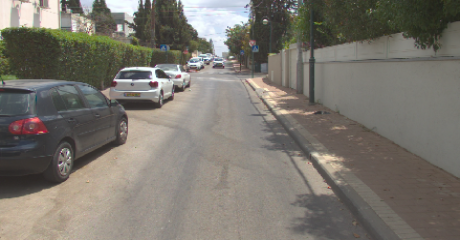} &
		\includegraphics[width=0.24\linewidth]{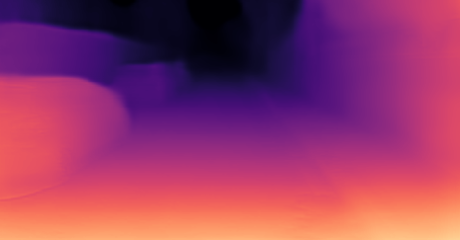} &
		\includegraphics[width=0.24\textwidth]{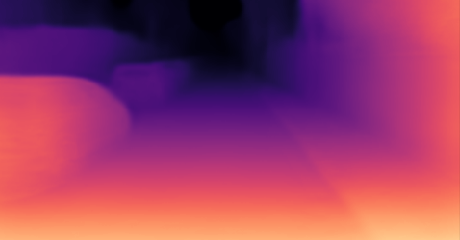} &
		\includegraphics[width=0.24\linewidth]{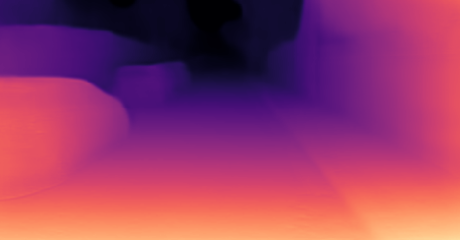}
		\\
		\includegraphics[width=0.24\linewidth]{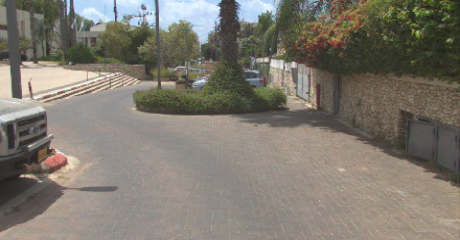} &
		\includegraphics[width=0.24\linewidth]{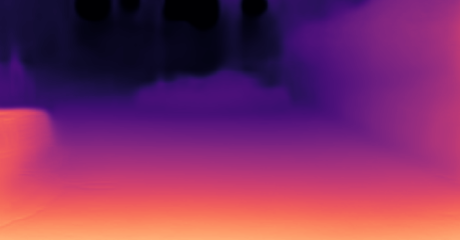} &
		\includegraphics[width=0.24\linewidth]{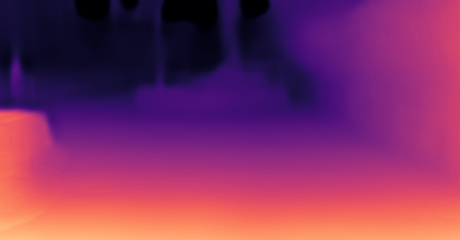} &
		\includegraphics[width=0.24\linewidth]{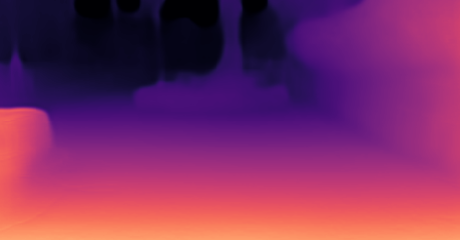} 
		\\
		\includegraphics[width=0.24\linewidth]{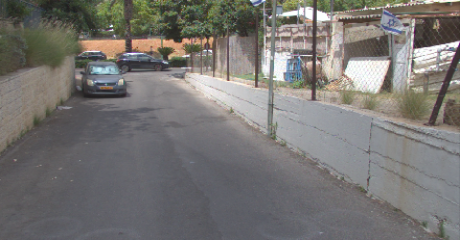} &
		\includegraphics[width=0.24\linewidth]{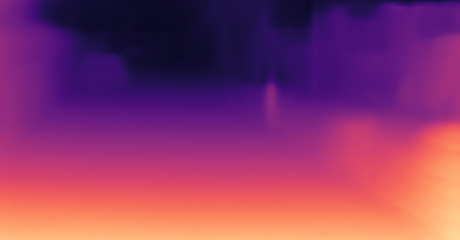} &
		\includegraphics[width=0.24\textwidth]{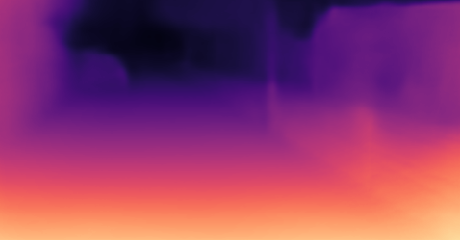} &
		\includegraphics[width=0.24\linewidth]{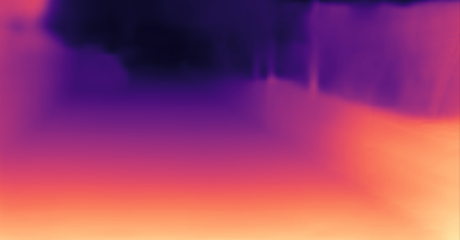}
		\\
		\includegraphics[width=0.24\linewidth]{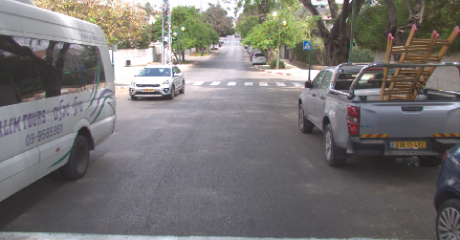} &
		\includegraphics[width=0.24\linewidth]{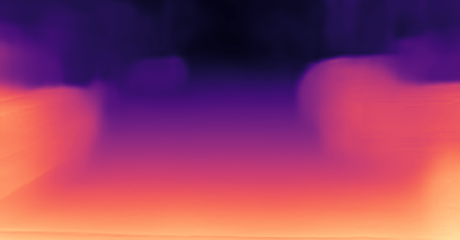} &
		\includegraphics[width=0.24\textwidth]{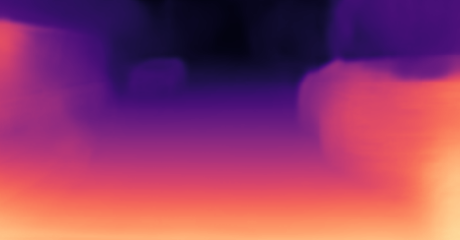} &
		\includegraphics[width=0.24\linewidth]{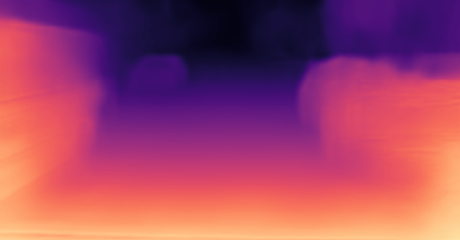}
		\\
		\includegraphics[width=0.24\linewidth]{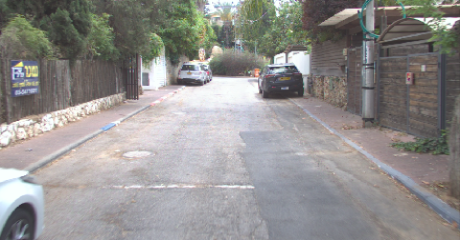} &
		\includegraphics[width=0.24\linewidth]{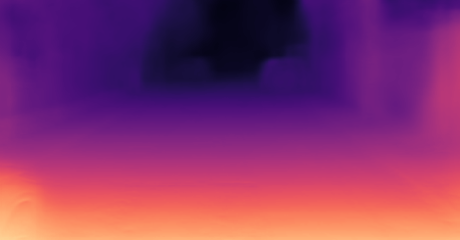} &
		\includegraphics[width=0.24\textwidth]{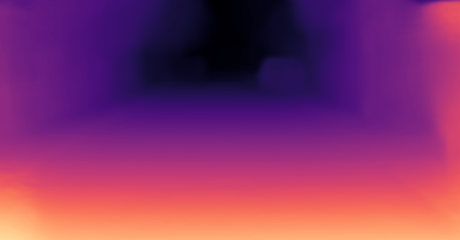} &
		\includegraphics[width=0.24\linewidth]{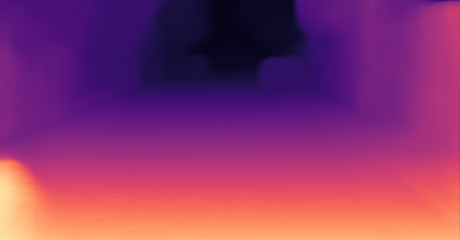}
		\\
		\includegraphics[width=0.24\linewidth]{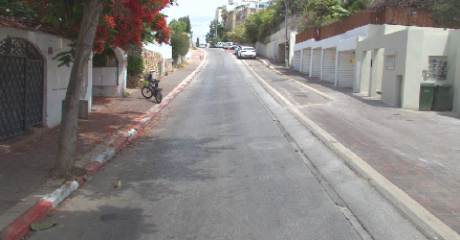} &
		\includegraphics[width=0.24\linewidth]{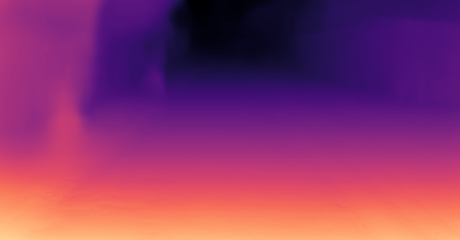} &
		\includegraphics[width=0.24\textwidth]{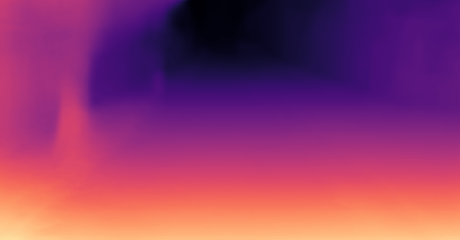} &
		\includegraphics[width=0.24\linewidth]{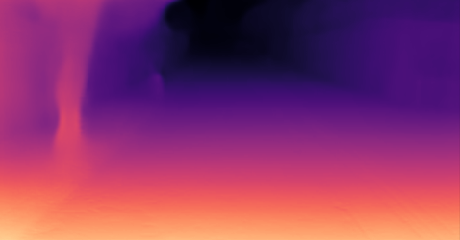}
		\\
		\includegraphics[width=0.24\linewidth]{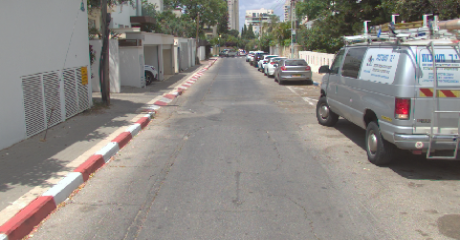} &
		\includegraphics[width=0.24\linewidth]{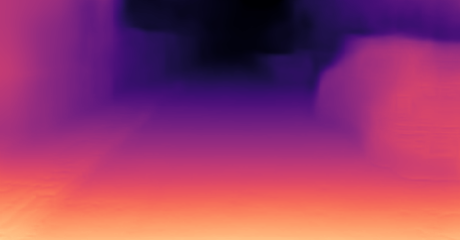} &
		\includegraphics[width=0.24\textwidth]{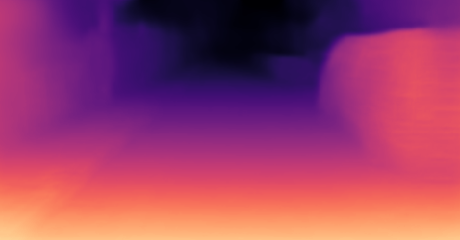} &
		\includegraphics[width=0.24\linewidth]{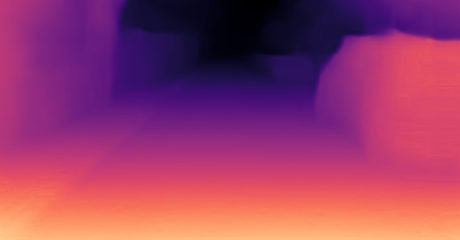}
		\\
		\includegraphics[width=0.24\linewidth]{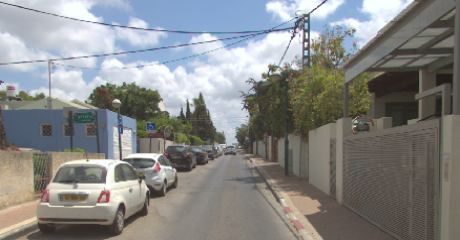} &
		\includegraphics[width=0.24\linewidth]{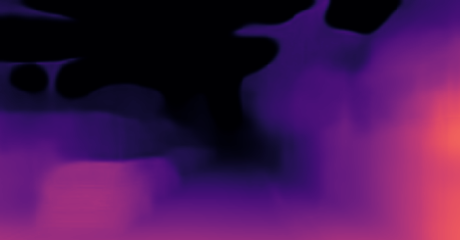} &
		\includegraphics[width=0.24\textwidth]{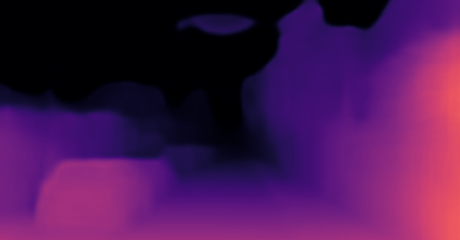} &
		\includegraphics[width=0.24\linewidth]{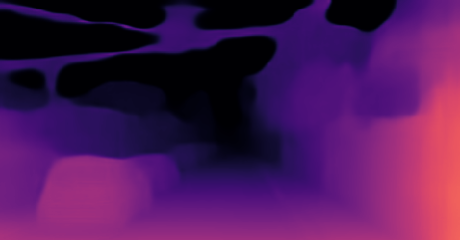}
	\end{tabular}
	\endgroup
	\caption{Qualitative results for the depth estimation task. pt-RGBP-Depth yields sharper edges and better recovers all structures. Best viewed zoomed in.}
	\label{fig:depth estimation qualitative results appendix}
\end{figure}

Figures \ref{fig:free space detection qualitative results appendix} and \ref{fig:depth estimation qualitative results appendix} present additional qualitative results of the free space detection and monocular depth estimation tasks respectively.

\end{document}